\documentclass{article}

\usepackage{microtype}
\usepackage{graphicx}
\usepackage{subcaption}
\usepackage{booktabs} % for professional tables
\usepackage{enumitem}

\usepackage{hyperref}

\usepackage[preprint]{icml2026}

\usepackage{amsmath}
\usepackage{amssymb}
\usepackage{mathtools}
\usepackage{amsthm}
\usepackage{multirow}

\usepackage[capitalize,noabbrev]{cleveref}

\theoremstyle{plain}

\theoremstyle{definition}

\theoremstyle{remark}

\usepackage[textsize=tiny]{todonotes}

\icmltitlerunning{Submission and Formatting Instructions for ICML 2026}

\begin{document}

\twocolumn[
  \icmltitle{A Decomposition Perspective to Long-context Reasoning for LLMs}

  \icmlsetsymbol{equal}{*}

  \begin{icmlauthorlist}
    \icmlauthor{Yanling Xiao}{tc}
    \icmlauthor{Huaibing Xie}{tc}
    \icmlauthor{Guoliang Zhao}{tc,sch}
    \icmlauthor{Shihan Dou}{tc,fd}
    \icmlauthor{Shaolei Wang}{tc}
    \icmlauthor{Yiting Liu}{tc}
    \icmlauthor{Nantao Zheng}{tct}
    \icmlauthor{Cheng Zhang}{tc}
    \icmlauthor{Pluto Zhou}{tc}
    \icmlauthor{Zhisong Zhang}{cityu}
    \icmlauthor{Lemao Liu}{fd}

  \end{icmlauthorlist}

  \icmlaffiliation{tc}{Large Language Model Department, Tencent}
  \icmlaffiliation{sch}{Xi'an Jiaotong University, Xi'an, China}
  \icmlaffiliation{tct}{Tencent}
  \icmlaffiliation{fd}{Fudan University, Shanghai, China}
  \icmlaffiliation{cityu}{City University of Hong Kong}

  \icmlcorrespondingauthor{Yanling Xiao}{lynniexiao@tencent.com}
  \icmlcorrespondingauthor{Pluto Zhou}{plutozhou096@foxmail.com}
  \icmlcorrespondingauthor{Lemao Liu}{lemaoliu@gmail.com}
  \icmlkeywords{Machine Learning, ICML}

  \vskip 0.3in
]

\printAffiliationsAndNotice{}

\newcommand{\lemao}[1]{\textcolor{orange}{\textbf{[By Lemao: #1]}}}




\begin{abstract}

Long-context reasoning is essential for complex real-world applications, yet remains a significant challenge for Large Language Models (LLMs). Despite the rapid evolution in long-context reasoning, current research often overlooks the internal complexity of the long-context reasoning task itself. In this paper, we move beyond this holistic view and decompose long-context reasoning into a set of fundamental atomic skills, and we then automatically synthesize a suite of pseudo datasets, each explicitly targeting a specific atomic skill. Our empirical analysis confirms that proficiency in these atomic skills is strongly correlated with general long-text reasoning performance. Building on this insight, we employ reinforcement learning on these pseudo datasets to sharpen the model's atomic skills, in the hope of boosting its general long-context reasoning ability. Extensive experiments across multiple benchmarks demonstrate the effectiveness of our approach: it outperforms a strong baseline by an average margin of 7.7\% (improving from 46.3\% to 54.0\%) across Loogle, Loong, LongBench-v2, BrowscompLong, Ruler-qa2, and MRCR. 

\end{abstract}

\section{Introduction}

%#########################

The rapid evolution of Large Language Models (LLMs) \cite{guo2025deepseek,liu2025deepseek,comanici2025gemini} has ushered in a new era of artificial intelligence, where the ability to handle extensive context windows is increasingly critical. From analyzing multi-document repositories to synthesizing legal contracts and financial reports~\cite{meyur2025benchmarking,reddy2024docfinqa}, real-world applications demand that the large language models not only ``comprehend'' massive amounts of text but also reason over them effectively. 
%While recent advancements have successfully scaled the context window size up to 1 million tokens\cite{xx}, a significant disparity remains between the {quantity} of context an LLM can ingest and the {quality} of reasoning it can perform\cite{xx}. 
Although recent advancements have expanded the maximum context window of LLMs to 1 million tokens \cite{team2024gemini,glm2024chatglm}, a pronounced chasm persists between the scale of context that models can accommodate and the efficacy of reasoning they can deliver \cite{paulsen2025context}.

% Existing studies often treat long-context reasoning as a monolithic skill, and they mainly employ fine-tuning or reinforcement learning algorithms to 

% primarily evaluating it through holistic benchmarks\cite{} or simple retrieval tasks like "Needle-in-a-Haystack" (NIAH)\cite{}. 

To bridge this critical gap, extensive research efforts have been dedicated to advancing the long-context reasoning capabilities of LLMs~\cite{chen2023longlora,li2024making,bai2024longalign}. Conventional approaches typically entail curating specialized training datasets tailored for long-context reasoning tasks \cite{chen2023longlora,zhang2025longreward,bai2024longalign}, followed by fine-tuning~\cite{li2024alr,zhang2025longreward} or reinforcement learning~\cite{QwenLongL1,wang2025loongrl} on these datasets to boost model performance. Nevertheless, long-context reasoning constitutes a monolithic and inherently complex task, rendering the direct construction of high-quality data fraught with challenges~\cite{yang2025longfaith}. Key challenges include the risk of misinformation stemming from inadequate verification protocols \cite{li2024large} and the potential for latent knowledge conflicts within curated datasets \cite{xu2024knowledge}.
%a surge of efforts has been made on enhancing the long-context resaoning capability for LLMs. Existing studies usually attempt to construct training data for long-context reasoning~\cite{xx}, and then perform fine-tuning or reinforcement learning on such data to enhance the model's long-context reasoning capabilities~\cite{xx}.
%However, long-context reasoning, as a monolithic task, is highly complex, and directly curating such data presents several challenges such as misinformation due to insufficient verification and potential knowledge conflicts~\cite{xx}.

% This paper proposes a decomposition perspective to re-examine long-text reasoning  rather than treating it as a monolithic skill. 

\begin{figure}
    \centering
    \includegraphics[width=0.9\linewidth]{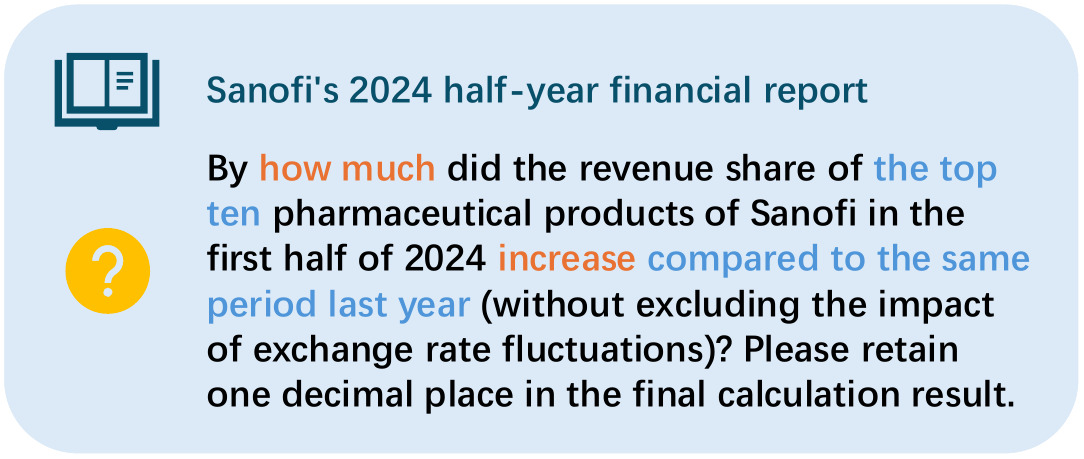}
    \caption{Decomposition of a complex task into atomic capabilities. The process necessitates Global Integration for aggregating distributed figures and Dynamic State Tracking for holding intermediate values during multi-step computation, rather than simple retrieval.}
    \label{fig:placeholder}
\end{figure}
In this paper, we propose a paradigm shift from a monolithic view of long-context reasoning to a \textbf{decomposition perspective}.
We argue that, from a cognitive standpoint, long-context reasoning is a hierarchical spectrum composed of fundamental \textit{atomic skills}. For instance, as illustrated in Figure 1, the task of calculating Sanofi's revenue share growth cannot be solved by simple retrieval. Instead, it necessitates \textit{Global Integration} to synthesize distributed financial data across different reporting periods (e.g., aggregating figures from H1 2024 and H1 2023), followed by \textit{Dynamic State Tracking} to execute multi-step reasoning—holding these intermediate values in memory to compute the final percentage increase. We decompose long-context reasoning into five atomic skills including \textit{Foundational Retrieval}, \textit{Anti-Interference}, \textit{Global Integration}, \textit{Relational Reasoning}, and \textit{Dynamic State Tracking} (\S 2.1). Unlike the complex long-context reasoning task, each atomic task is comparatively straightforward; thus, we can relatively easily curate training data for each atomic skill through an anchor-based automatic pipeline with verification (\S 2.2). Our empirical experiments further demonstrate that these atomic skills are strongly correlated with long-context reasoning skill (\S 3). This finding indicates that enhancing these atomic skills of LLMs can ultimately boost their long-text reasoning performance.

Based on the curated datasets for these atomic skills, we then present a highly efficient training strategy: we employ Reinforcement Learning (RL)~\cite{shao2024deepseekmath,yu2025dapo} in the curated set of approximately 4,000 synthetic samples generated through our pipeline. This targeted approach sharpens the model's atomic capabilities, enabling it to generalize to complex, unseen long-context reasoning tasks. Experimental results on six challenging benchmarks—including Loogle~\cite{li2024loogle}, Loong~\cite{wang2024leave} and LongBench v2~\cite{bai2025longbench}—show that our approach significantly improves performance, outperforming strong baselines such as DeepSeek-R1-distill-32B~\cite{DeepSeekR1Distill32B} by an average margin of 7.7\% (improved from 46.3\% to 54.0\%) (\S 4).

Our contributions are summarized as follows:
\begin{itemize}

    \item \textbf{Taxonomy of Atomic Skills:} We decompose the long-context reasoning into five distinct, hierarchical capabilities, providing a anchor-based pipeline to automatically synthesize data for these atomic skills (\S 2).
    % a granular analysis for understanding and evaluating these atomic skills for LLMs.
    %\item \textbf{Scalable Data Synthesis:} We propose the Anchor-based Reasoning (AbR) framework, an automated pipeline for generating controllable, verifiable, and infinitely scalable training data targeting specific cognitive demands.
    \item \textbf{Validation via Correlation:} We provide an empirical evidence that our proposed atomic skills are statistically correlated well with the general long-context reasoning capability (\S 3).
    \item \textbf{Efficient RL-based Enhancement:} We demonstrate that targeted Reinforcement Learning on a small scale ($4k$) of atomic-skill data yields substantial improvements in general long-context reasoning, establishing a data-efficient path for model alignment (\S 4).  
\end{itemize}

%#########################
\section{Atomic Skills for Long-context Reasoning}

In essence, long-context reasoning is not a monolithic skill but a hierarchical spectrum of cognitive requirements. Therefore, we decompose long-context reasoning into five atomic skills, ordered by increasing cognitive complexity: from foundational retrieval to robust discrimination, global aggregation, rational reasoning, and finally, dynamic state manipulation. 

\subsection{Atomic Skill Taxonomy}
\paragraph{Foundational Retrieval: Needle-in-a-Haystack (NIAH)}
The hierarchy begins with Foundational Retrieval, the most fundamental skill. Before any complex reasoning can occur, a model must first prove it can reliably locate a specific piece of information (``the needle'') anywhere within a vast sea of text (``the haystack''), overcoming the common ``lost-in-the-middle'' problem. This is the bedrock of all long-context capabilities.

\paragraph{Robustness to Noise: Anti-Interference Capability}
Building on simple retrieval, Robustness to Noise addresses a more realistic challenge. It’s not enough to just find information; a model must distinguish the correct target from similar-looking but incorrect ``distractors''. This skill measures the ability to maintain focus and factual accuracy when faced with deceptive or confusing information.

\paragraph{Global Integration: Multi-Source Information Processing}
Moving beyond finding a single, correct piece of evidence, Global Integration requires a model to locate and synthesize information from multiple, separate locations within the context. Instead of retrieving one fact, the model must connect several scattered data points to construct a single, coherent answer, demonstrating an ability to process information in parallel.

\paragraph{Relational Reasoning: Structure Understanding and Logic} 
The next level of complexity, Relational Reasoning, requires more than just gathering facts; it demands an understanding of the logical relationships between them. A model must recognize the text's underlying structure to perform operations like filtering, joining, or comparing different sets of information, much like executing a database query on unstructured text.

\paragraph{Dynamic State Tracking: Long-Range Computational Reasoning} 
At the peak of this hierarchy, Dynamic State Tracking tests a model's ability to perform multi-step computational reasoning. Here, a model must not only retrieve and relate information but also use it to perform intermediate calculations. It must derive new values from the text, hold these ``states'' in its working memory, and then execute a final computation using these derived results, completing a full ``retrieve-solve-then-compute'' workflow.

\subsection{Automated Dataset Construction Pipeline}

\begin{figure}
    \centering
    \includegraphics[width=0.9\linewidth]{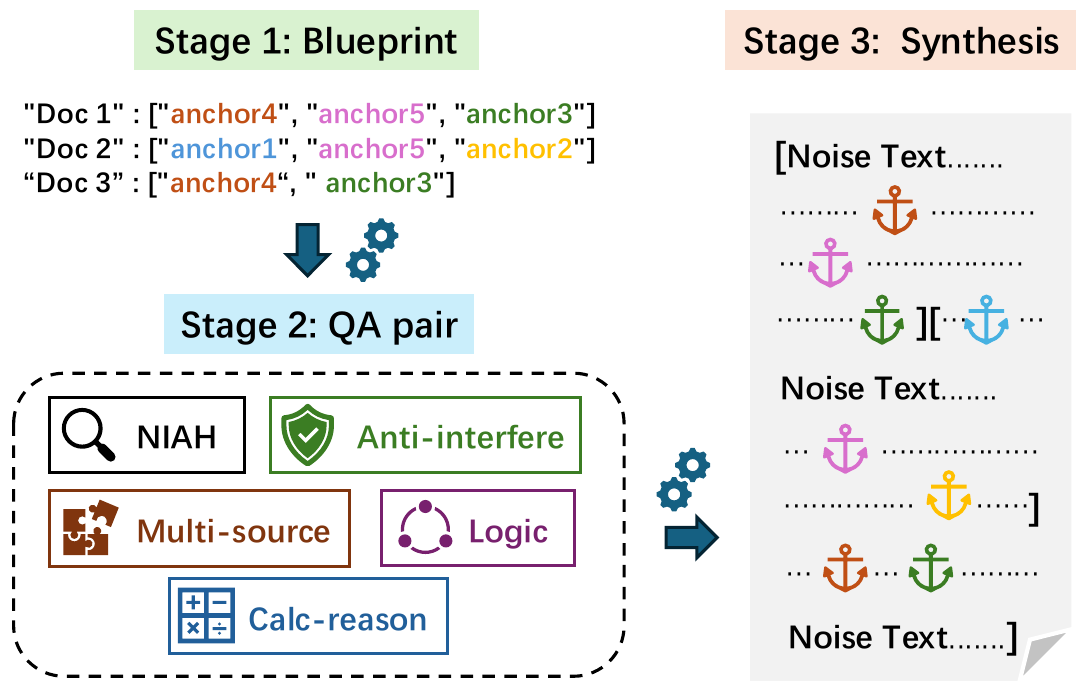}
    \caption{The Automated Dataset Construction Pipeline of the Anchor-based Reasoning (AbR) Framework.  }
    \label{fig:overview}
\end{figure}

To systematically evaluate the hierarchical cognitive demands outlined in our taxonomy, we introduce the \textbf{Anchor-based Reasoning(AbR)} framework. The core principle of this framework is to embed algorithmically generated anchors—unique strings paired with specific, verifiable questions—into extensive, noise-laden documents. By strategically distributing these ``anchor-question'' pairs, we can then pose meta-questions that require a model to aggregate or reason over the answers to these individual embedded queries. This design transforms the ambiguous challenge of long-context reasoning into a precise, quantifiable workflow comprising three core steps: information localization, embedded problem-solving, and logical integration. As shown in figure~\ref{fig:overview}, we establish a three-stage automated pipeline to construct scalable, controllable, and verifiable datasets.
% To implement this design and construct a dataset that is infinitely scalable, controllably difficult, and automatically verifiable, we establish the following three-stage automated pipeline.

\paragraph{Stage 1: Logical Blueprint Generation.}
We first programmatically generate a \texttt{doc\_mapping} JSON file. This blueprint defines the ``ground truth'' by specifying which anchors (\texttt{anc\_id}) are contained within each virtual document (\texttt{doc\_id}).

\begin{verbatim}
// 5 anchors across 3 docs.
{
  "doc1": ["anc4", "anc5", "anc3"],
  "doc2": ["anc1", "anc5", "anc2"],
  "doc3": ["anc4", "anc3"]
}
\end{verbatim}

\paragraph{Stage 2: Question-Answer Pair Generation.}
Based on \texttt{doc\_mapping}, we then generate diverse meta-questions requiring complex reasoning via rule-based templates or LLMs.
The answers are generated as an \textit{executable expression}, representing the procedural steps needed to arrive at the correct solution. $Solve(Q(anc))$ denotes solving the question associated with a specific anchor.
\begin{itemize}
    \item \textit{Question (Relational):} ``In the document that contains both `anchor5' and `anchor3', what is the answer to the question associated with `anchor4'?''
    % \item \textit{Answer Expression:} $solve(Q(anchor4, in\_doc(intersect(docs(anchor5), docs(anchor3)))))$
    \item $\text{solve}\left(Q\left(\text{anc}_4, \text{in\_doc}\left(\text{docs}(\text{anc}_5) \cap \text{docs}(\text{anc}_3)\right)\right)\right)$
    \item \textit{Question (Computational):} ``Calculate the sum of the answers to the questions for `anchor3' in all documents that contain it.''
    % \item \textit{Answer Expression:} $sum([solve(Q(doc1, key3)), solve(Q(doc3, key3))])$
    \item $\sum \left[ \text{solve}(Q(\text{doc}_1, \text{anc}_3)), \text{solve}(Q(\text{doc}_3, \text{anc}_3)) \right]$
\end{itemize}

\paragraph{Stage 3: Multi-Document Context Synthesis.}
We assemble the final sample by: 1) selecting unrelated background texts for each \texttt{doc\_id}; 2) inserting anchor-question pairs at random positions based on the \texttt{doc\_mapping}; and 3) pairing the meta-question with the full synthesized context.

\paragraph{Skill-Specific Task Construction} We tailor AbR tasks for each atomic skill: (1) \textit{Foundational Retrieval} inserts specific pairs requiring the model to locate distributed anchors for objective answers. (2) \textit{Robustness to Noise} employs two interference patterns: \textit{Similarity Discrimination} uses highly similar anchors for fine-grained distinction, while \textit{Conflicting Information Resolution} distributes identical anchors to enforce contradiction resolution. (3) \textit{Global Integration} fragments clues across separate documents, compelling the model to aggregate dispersed data points into coherent logical chains. (4) \textit{Relational Reasoning} imposes logical constraints on structural positions, requiring set operations (e.g., intersection, union) on document locations to identify targets. (5) \textit{Dynamic State Tracking} necessitates a multi-stage process deriving numerical values from distributed anchors to execute sequential mathematical operations. Detailed showcases for each skill are provided in the Appendix ~\ref{showcases}.

\paragraph{Controllable Difficulty and Curriculum} A key advantage of our methodology is the precise control of training difficulty across a continuous complexity spectrum. By systematically tuning parameters such as context length, anchor density, noise similarity, and reasoning depth, we establish a controlled environment for generating diverse challenges. This granularity facilitates the creation of a fine-grained training curriculum, enabling models to progressively advance their long-context capabilities.

\section{Validating the Role of Atomic Skills in Long-context Reasoning}

With the AbR pipeline enabling the precise generation of datasets targeting specific atomic skills, we proceed to verify the ecological validity of our taxonomy. We aim to confirm that the atomic skills are not merely theoretical constructs but foundational drivers of performance in complex, real-world scenarios. By conducting rigorous analyses, we demonstrate that the proposed atomic skills serve as critical indicators of the model's overall capability.

\subsection{Setup}
% We evaluated 11 LLMs on both real-world long-context benchmarks and our proposed atomic skill evaluation sets. To quantify the relationship between atomic skill proficiency and long-context reasoning, we employed the Spearman rank correlation coefficient ($\rho$). All evaluations were conducted with context lengths up to 128K tokens to maintain experimental consistency.
% \begin{enumerate}[leftmargin=*]
%     \item \textbf{Target Skill (Long-context Reasoning):} We use several standard long-context benchmarks, such as Loogle~\cite{li-etal-2024-loogle}, LongBench-v2~\cite{bai-etal-2024-longbenchv2} and Loong~\cite{wang-etal-2024-leave}, to measure models' general long-context capability.
%     \item \textbf{Atomic Skills (Our Taxonomy):} We constructed five evaluation sets based on our proposed atomic skills: \textit{Needle-in-a-Haystack (NIAH)}, \textit{Anti-Interference}, \textit{Multi-Source}, \textit{Logic}, and \textit{Calc\_Reason}.
%     \item \textbf{Atomic Skill Control Group:} We additionally introduced existing open-source benchmarks focusing on computational reasoning and information aggregation capabilities as a control group.
%     \item \textbf{LLMs:} We selected 11 open-source models with parameter sizes ranging from 7B to 32B to evaluate whether standard atomic skill tasks correlate well with the target skill across different model capacities. ~\footnote{We employ identical inference parameters across all models, utilizing YaRN for length extrapolation on those with context windows smaller than 128K.}
% \end{enumerate}

In these analyses, we evaluate LLMs with both standard real-world long-context benchmarks and atomic skill evaluation sets. The real-world benchmarks are used to measure models' general long-context capability, including Loogle~\cite{li2024loogle}, LongBench-v2~\cite{bai2025longbench} and Loong~\cite{wang2024leave}. The atomic evaluation sets are based on our proposed atomic skills: \textit{Needle-in-a-Haystack (NIAH)}, \textit{Anti-Interference}, \textit{Multi-Source}, \textit{Logic}, and \textit{Calc\_Reason}. We additionally introduced existing open-source benchmarks focusing on computational reasoning and information aggregation capabilities as a control group.

We selected 11 open-source models with parameter sizes ranging from 7B to 32B. To quantify the relationship between atomic skill proficiency and long-context reasoning, we employed the Spearman rank correlation coefficient ($\rho$). All evaluations were conducted with context lengths up to 128K tokens to maintain experimental consistency.

\subsection{Analysis 1: Correlation Analysis}
% Figure \ref{fig:correlation_heatmap} illustrates the Spearman correlation coefficients. The results provide strong statistical evidence ($\rho$-values) validating our decomposition of long-context intelligence.
\begin{figure}[t]
    \centering
    \includegraphics[width=0.97\linewidth]{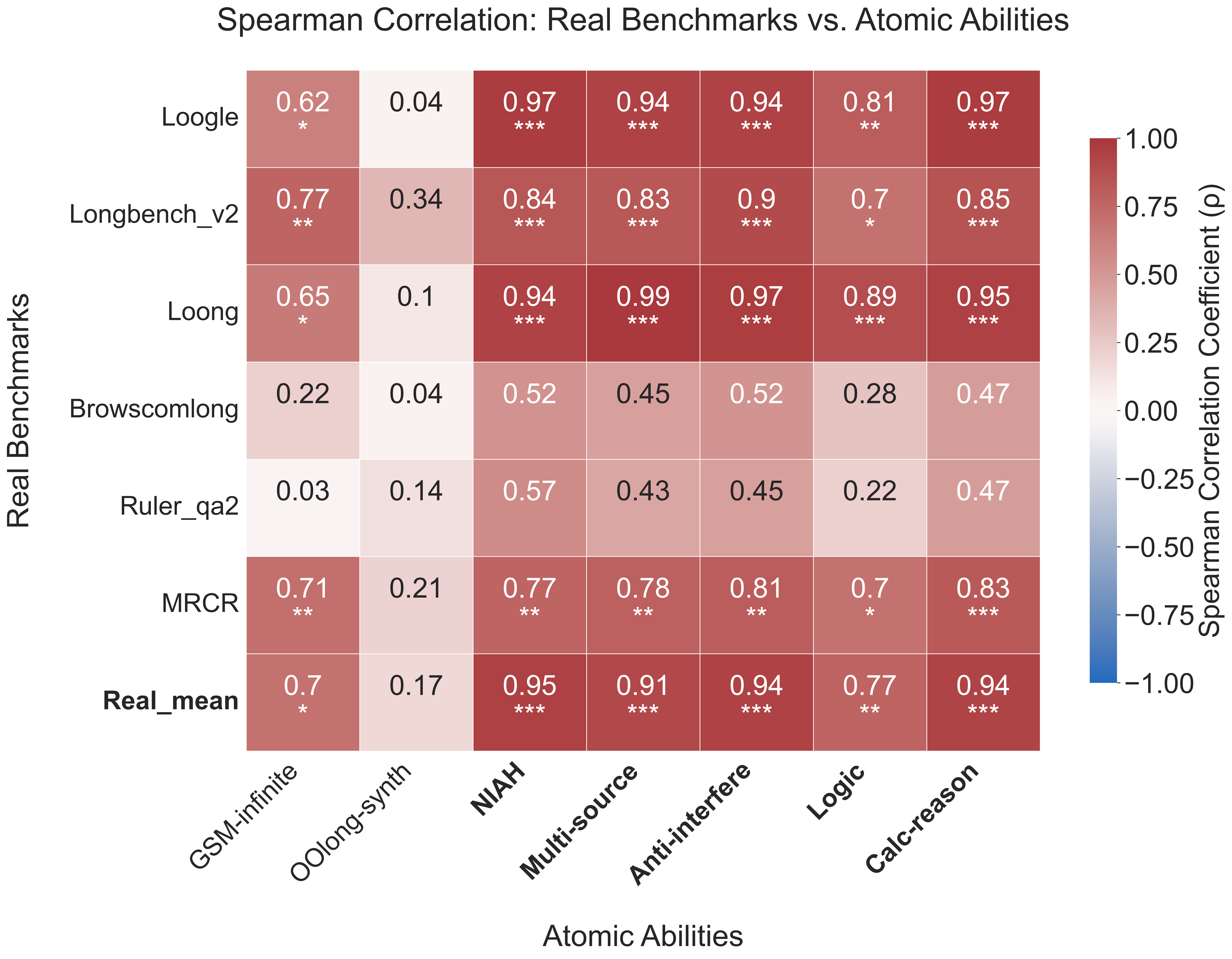}
    \caption{\textbf{Spearman Correlation Analysis.} The heatmap compares the correlation of our proposed atomic capabilities against real-world long-context benchmarks. }
    \label{fig:correlation_heatmap}
\end{figure}
We first analyze the Spearman correlation coefficients between the performance on existing real-world and proposed atomic benchmarks. As shown in Figure \ref{fig:correlation_heatmap}, the results provide strong statistical evidence ($\rho$-values), validating the effectiveness of our approach.

\paragraph{High Predictive Validity.} 
Our atomic probes demonstrate exceptional predictive power regarding the average performance on real-world benchmarks (\textit{Real\_mean}). For instance, our \textit{NIAH} and \textit{Anti-interfere} probes achieve exceptional alignment with \textit{Real\_mean} ($\rho=0.95$ and $\rho=0.94$, respectively). This superiority is particularly evident in challenging benchmarks like \textit{Loong}, where \textit{Multi-source} reaches a correlation of $\rho=0.99$. These consistently high correlations (all significant at $p<0.001$) validate our taxonomy: rather than being an arbitrary collection of tasks, these probes serve as accurate ``proxies'' that effectively decompose the complexity of long-context understanding into measurable atomic units.

\paragraph{Inadequacy of Generic Baselines.} 
The results further show that generic baselines exhibit limited predictive power for real-world performance. The synthetic baseline \textit{OOlong-synth} shows a negligible correlation with the average of real benchmarks (\textit{Real\_mean}, $\rho=0.17$). While \textit{GSM-infinite} demonstrates a moderate correlation ($\rho=0.70$), it consistently lags behind \textit{Calc-reason} ($\rho=0.94$).

\subsection{Analysis 2: Diagnosing the Capability Gap}
\begin{table*}[t]
\centering
\caption{Performance comparison on atomic skills.}
\label{tab:atomic_capability_probes}
%\resizebox{\textwidth}{!}
{%
\begin{tabular}{lrrrrr}
        \toprule
        \textbf{Model} & \textbf{NIAH} & \textbf{Anti-interfere} & \textbf{Multi\_source} & \textbf{Logic} & \textbf{Calc\_reason} \\
        \midrule
        QwenLong-L1-32B & 66.50\% & 33.83\% & 38.81\% & 28.50\% & 47.37\% \\
        Qwen2.5-32b-instruct & 37.00\% & 22.13\% & 23.88\% & 13.00\% & 36.59\% \\
        DeepSeek-R1-Distill-Qwen-32B & 58.25\% & 25.96\% & 32.54\% & 17.50\% & 42.11\% \\
        Qwen3-32B & 23.00\% & 12.55\% & 23.28\% & 19.00\% & 19.80\% \\
        Qwen3-32B & 69.50\% & 29.15\% & 37.01\% & 27.25\% & 52.13\% \\
        Qwen3-30B-A3B-thinking-2507 & 74.25\% & 41.70\% & 47.46\% & 31.50\% & 60.15\% \\
        Qwen3-14B & 46.50\% & 21.06\% & 29.25\% & 19.50\% & 37.09\% \\
        Deepseek-R1-Distill-Qwen-14B & 37.50\% & 13.40\% & 19.10\% & 9.00\% & 24.31\% \\
        Qwen2.5-14b-instruct & 27.75\% & 14.47\% & 19.70\% & 14.25\% & 31.58\% \\
        Qwen3-8B & 42.00\% & 17.66\% & 25.37\% & 15.00\% & 37.09\% \\
        Qwen2.5-7b-instruct & 16.50\% & 8.30\% & 11.64\% & 9.25\% & 19.30\% \\
        Deepseek-R1-Distill-Qwen-7B & 5.25\% & 2.13\% & 3.28\% & 3.75\% & 4.51\% \\
        \bottomrule
    \end{tabular}%
}
\end{table*}

By cross-referencing the correlation data (Figure \ref{fig:correlation_heatmap}) with the absolute performance scores (Table \ref{tab:atomic_capability_probes}), we further identify critical structural flaws in current models.

% \paragraph{The ``Importance-Proficiency'' Mismatch.}
% There is a stark contrast between what matters and what models can do.
% \begin{itemize}
%     \item \textbf{High Correlation, Low Performance:} The heatmap shows that \textbf{Anti-interfere} and \textbf{Multi-source} are strong predictors of real-world success ($\rho=0.94$ and $\rho=0.91$, respectively). However, Table \ref{tab:atomic_capability_probes} shows that models struggle significantly on these specific tasks. For example, while \textit{Qwen2.5-32b-instruct} achieves 37.00\% on basic NIAH, it drops to \textbf{22.13\%} on Anti-interfere and \textbf{23.88\%} on Multi-source. Similarly, \textit{Logic} proves to be the most difficult task (e.g., 13.00\% for the same model), though its correlation with real-world performance is moderate ($\rho=0.77$).
%     \item \textbf{The Retrieval Ceiling:} Most models perform relatively well on clean \textit{NIAH} tasks. However, since NIAH is merely a prerequisite, improving it further yields diminishing returns for complex real-world tasks.
% \end{itemize}

\paragraph{The ``Importance-Proficiency'' Mismatch.}
Our analysis reveals a specific pattern of High Correlation, Low Performance. For example, \textit{Anti-interfere} and \textit{Multi-source} are strong predictors of real-world success ($\rho=0.94$ and $\rho=0.91$, respectively). However, most models struggle on these tasks: while \texttt{Qwen2.5-32b-instruct} achieves 37.00\% on basic NIAH, it drops to \textbf{22.13\%} on Anti-interfere and \textbf{23.88\%} on Multi-source. Similarly, \textit{Logic} proves to be the most difficult task (e.g., 13.00\% for the same model), though its correlation with real-world performance is moderate ($\rho=0.77$). Moreover, there exists a Retrieval Ceiling, that is, most models perform relatively well on clean \textit{NIAH} tasks. However, since NIAH is merely a prerequisite, improving it further yields diminishing returns for complex real-world tasks.

\paragraph{The Robustness Bottleneck.}
The perform difference between \textit{NIAH} and \textit{Anti-Interference} (e.g., 37.00\% v.s. 22.13\% for \texttt{Qwen2.5-32b-instruct}) highlights that models lack discrimination capabilities. They can retrieve information but tend to be easily distracted by ``lure'' noise. Since \textit{Anti-Interference} correlates highly with real-world benchmarks ($\rho=0.94$), this fragility can be an important factor for general performance.

\section{RL-Based Enhancement}
\subsection{Setup}

\paragraph{Training Setup}
To enhance atomic skills, we employ the Group Relative Policy Optimization (GRPO) algorithm~\cite{shao2024deepseekmath}. 
% Unlike standard PPO, GRPO eliminates the critic network by computing the advantage for each output relative to the average reward of a group. 
To mitigate reward homogenization and accelerate convergence, we incorporate Dynamic Sampling~\cite{yu2025dapo} to prune redundant trajectories. The reward signal is derived from an LLM-as-a-Judge framework using \texttt{gpt-oss-120B}~\cite{agarwal2025gpt}, which assigns binary correctness rewards. Specifically for instruction-tuned models, we introduce a Chain-of-Thought (CoT) system prompt and a format-compliance reward to induce deliberate reasoning, whereas models with CoT follow standard procedures. More training details are showned in Appendix~\ref{training_details}.

We conduct our main experiments using three backbone models: \texttt{Qwen2.5-14B-Instruct}, \texttt{Qwen2.5-32B-Instruct}~\cite{Qwen2.5}, and \texttt{DeepSeek-R1-Distill-32B}~\cite{DeepSeekR1Distill32B}, with ablation studies and in-depth analyses performed on the latter. All models are trained from scratch (cold-started) without prior fine-tuned policies. We construct the training dataset by sampling DeepSeek V3.1, filtering for queries with a pass rate between 0.3 and 0.6 to ensure appropriate difficulty. The final dataset mixture adheres to the ratio of Anti-interfere : Multi-hop : Multi-source : Logic : Calc-reason : NIAH = $5:3:2:2:2:1$.
During the rollout phase, we set sampling hyperparameters to $top\_p=0.6$ and $top\_k=20$. The maximum sequence lengths for input and output are restricted to 24k and 8k tokens, respectively. All models are trained on 64 H20 GPUs.

% \paragraph{Training Setup}
% We conduct our main experiments using three backbone models: \texttt{Qwen2.5-14B-Instruct}, \texttt{Qwen2.5-32B-Instruct}~\cite{Qwen2.5}, and \texttt{DeepSeek-R1-Distill-32B}~\cite{DeepSeekR1Distill32B}. Ablation studies and in-depth analyses are performed on the \texttt{DeepSeek-R1-Distill-32B} model. We employ the Group Relative Policy Optimization (GRPO) algorithm for training. During the rollout phase of training, we set the sampling hyperparameters to $top\_p=0.6$ and $top\_k=20$. The maximum sequence lengths for input and output during these rollouts are restricted to 24k and 8k tokens, respectively. We utilize \texttt{gpt-oss-120B} as the underlying reward model. Specifically for the instruction-tuned models, we introduce a specialized ``slow-thinking'' system prompt alongside a format-compliance reward signal to guide the models toward a deliberate reasoning mode. All models are trained on 64 H20 GPUs.

\paragraph{Evaluation}
Our evaluation framework incorporates both standard open-source benchmarks and a custom atomic capability dataset. For open-source benchmarks, we select \texttt{LongBench-v2}~\cite{bai2025longbench}, \texttt{Loong}~\cite{wang2024leave}, \texttt{MRCR}~\cite{MRCR}, \texttt{BrowsCompLong}~\cite{BrowseComplong}, the \texttt{qa\_2} subset from \texttt{Ruler}~\cite{hsieh2024ruler}, and the real-prompt subset of \texttt{Loogle}~\cite{li2024loogle}. To ensure consistency, we filter these datasets to include only samples with context lengths less than or equal to 128k tokens. The evaluation metrics for these open-source benchmarks align strictly with the protocols defined in their respective original papers. Additionally, we assess atomic capabilities using a dataset constructed via our proposed methodology, employing \texttt{gpt-oss-120B}~\cite{GPTOSS120B} to conduct consistency-based evaluation.

\paragraph{Baselines}
We compare our approach against some general baselines including the closed-source (\texttt{Gemini-3-Pro}) as well as open-source baselines including \texttt{Kimi-K2-Thinking}~\cite{team2025kimi}, \texttt{DeepSeek-V3.1}~\cite{DeepSeekV3.1}, \texttt{QwenLong-L1-32B}~\cite{QwenLongL1}, and \texttt{Qwen3-235B}~\cite{yang2025qwen3}. In addition, to further show the superiority of our approach, we compare it with some direct baselines, which are obtained by training DeepSeek-R1-distill-32B on the synthetic long-context reasoning datasets named LongReason~\cite{ling2025longreason}, QwenDocqa~\cite{QwenLongL1}, LoonngRl~\cite{wang2025loongrl}. In  details, we utilize the officially released datasets for LongReason and QwenDocQA. Additionally, we reproduce the data construction pipeline of Loonngrl to generate a dataset of $4,000$ samples for comparison.
% We compare our approach against strong proprietary and open-source baselines. For open-source models, we include \texttt{DeepSeek-V3.1}~\cite{DeepSeekV3.1} and two versions (L1 and L1.5) of \texttt{Qwen-Long}~\cite{QwenLong,QwenLong1.5} . 
To accommodate the long-context requirements of our evaluation, for any baseline model with a native context window smaller than 128k, we apply YaRN for length extrapolation.

\subsection{Main Results}

\label{sec:main_results}

\begin{table*}[t]
\centering
\caption{Performance comparison on Open Long Context Benchmarks. The top block shows general baselines and the bottom block illustrates the direct baselines with the same backbone. The best result in each column within its block is in \textbf{bold}. 
%Within each Qwen comparison block, the better result is also highlighted in \textbf{bold}.
}
\label{tab:main_results}
\resizebox{\textwidth}{!}{
\begin{tabular}{l rrrrrrr}
\toprule
\textbf{Model} & \textbf{Loogle} & \textbf{Longbench-v2} & \textbf{Loong} & \textbf{Browscomplong} & \textbf{Ruler-qa2} & \textbf{MRCR} & \textbf{Average} \\
\midrule
% --- Baselines ---
Gemini-3-pro & 52.86\% & \textbf{69.38\%} & \textbf{65.43\%} & \textbf{88.07\%} & \textbf{83.01\%} & \textbf{75.30\%} & \textbf{72.34\%} \\
Kimi-K2-Thinking & 51.50\% & 49.30\% & 58.01\% & 58.10\% & 49.98\% & 51.77\% & 53.11\% \\
DeepSeek-V3.1 & \textbf{55.77\%} & 52.88\% & 50.55\% & 56.27\% & 42.97\% & 46.62\% & 50.84\% \\
Qwen3-235B-A22B-thinking-2507 & 52.77\% & 49.30\% & 53.70\% & 50.76\% & 46.81\% & 44.61\% & 49.66\% \\
QwenLong-L1-32B & 49.32\% & 43.74\% & 44.68\% & 69.93\% & 47.70\% & 27.70\% & 47.18\% \\
\midrule
DeepSeek-R1-distill-32B & 42.31\% & 43.94\% & 38.17\% & 64.22\% & 57.23\% & 31.94\% & 46.30\% \\
DeepSeek-R1-distill-32B+LongReason & 43.23\% & 44.14\% & 38.57\% & 57.90\% & 58.36\% & 32.85\% & 45.84\% \\
DeepSeek-R1-distill-32B+QwenDocqa & 47.96\% & 46.32\% & 39.98\% & 71.56\% & 67.45\% & 34.59\% & 51.31\% \\
DeepSeek-R1-distill-32B+LoongRL & 48.41\% & 45.73\% & 40.29\% & 70.74\% & 61.84\% & \textbf{37.23\%} & 50.71\% \\
\textbf{DeepSeek-R1-distill-32B+Ours}& 50.59\% & 49.70\% & \textbf{44.68\%} & \textbf{73.09\%} & \textbf{69.38\%} & 36.74\% & 54.03\% \\
\textbf{DeepSeek-R1-distill-32B+Ours+LoongRL} & \textbf{55.59\%} & \textbf{51.29\%} & 44.45\% & 72.27\% & 67.01\% & 36.12\% & \textbf{54.46\%} \\
% % --- Qwen2.5-14B ---
% Qwen2.5-14B-instruct & 33.06\% & 36.38\% & 24.65\% & 47.30\% & 40.28\% & 31.87\% & 35.59\% \\
% Qwen2.5-14B-instruct+Ours & \textbf{40.51\%} & \textbf{41.15\%} & \textbf{28.63\%} & \textbf{69.72\%} & \textbf{63.21\%} & 31.73\% & \textbf{45.83\%} \\
% \midrule
% % --- Qwen2.5-32B ---
% Qwen2.5-32B-instruct & 35.97\% & 39.76\% & 34.27\% & 60.65\% & 45.25\% & 33.33\% & 41.54\% \\
% Qwen2.5-32B-instruct+Ours & \textbf{45.78\%} & \textbf{43.94\%} & \textbf{38.34\%} & \textbf{60.86\%} & \textbf{47.74\%} & \textbf{36.33\%} & \textbf{45.50\%} \\
\bottomrule
\end{tabular}
}
\end{table*}

\paragraph{Overall Performance on Long Context Benchmarks}
Table \ref{tab:main_results} presents the comprehensive evaluation results across six challenging long-context benchmarks. We can see that our approach yields an absolute gain of 7.7\% in average over the backbone \texttt{DeepSeek-R1-distill-32B} and it consistently outperforms all three direct baselines including LongReason, QwenDocqa, LoogRL. Moreover,  
Our approach achieves superior performance over the remarkable \texttt{Kimi-K2-Thinking} model, outperforming \texttt{DeepSeek-V3.1} and \texttt{Qwen3-235B} by a large margin.

\begin{table*}[t]
\centering
\caption{Performance of our approach applied to more backbone models. 
The best result in each column with its block is highlighted in \textbf{bold}.}
\label{tab:comparison_results}
\resizebox{\textwidth}{!}{
\begin{tabular}{l rrrrrrr}
\toprule
\textbf{Model} & \textbf{Loogle} & \textbf{Longbench-v2} & \textbf{Loong} & \textbf{Browscomplong} & \textbf{Ruler-qa2} & \textbf{MRCR} & \textbf{Average} \\
\midrule
% --- Ablation on DeepSeek-R1-distill-32B ---
% DeepSeek-R1-distill-32B & 42.31\% & 43.94\% & 38.17\% & 64.22\% & 57.23\% & 31.94\% & 46.30\% \\
% DeepSeek-R1-distill-32B+LongReason & 43.23\% & 44.14\% & 38.57\% & 57.90\% & 58.36\% & 32.85\% & 45.84\% \\
% DeepSeek-R1-distill-32B+QwenDocqa & 47.96\% & 46.32\% & 39.98\% & 71.56\% & 67.45\% & 34.59\% & 51.31\% \\
% DeepSeek-R1-distill-32B+Loongrl & 48.41\% & 45.73\% & 40.29\% & 70.74\% & 61.84\% & \textbf{37.23\%} & 50.71\% \\
% DeepSeek-R1-distill-32B+Ours & 50.59\% & 49.70\% & \textbf{44.68\%} & \textbf{73.09\%} & \textbf{69.38\%} & 36.74\% & 54.03\% \\
% DeepSeek-R1-distill-32B+Ours+Loongrl & \textbf{55.59\%} & \textbf{51.29\%} & 44.45\% & 72.27\% & 67.01\% & 36.12\% & \textbf{54.46\%} \\

% --- Qwen2.5-14B ---
Qwen2.5-14B-instruct & 33.06\% & 36.38\% & 24.65\% & 47.30\% & 40.28\% & 31.87\% & 35.59\% \\
Qwen2.5-14B-instruct+Ours & \textbf{40.51\%} & \textbf{41.15\%} & \textbf{28.63\%} & \textbf{69.72\%} & \textbf{63.21\%} & 31.73\% & \textbf{45.83\%} \\
\midrule
% --- Qwen2.5-32B ---
Qwen2.5-32B-instruct & 35.97\% & 39.76\% & 34.27\% & 60.65\% & 45.25\% & 33.33\% & 41.54\% \\
Qwen2.5-32B-instruct+Ours & \textbf{45.78\%} & \textbf{43.94\%} & \textbf{38.34\%} & \textbf{60.86\%} & \textbf{47.74\%} & \textbf{36.33\%} & \textbf{45.50\%} \\
\bottomrule
\end{tabular}
}
\end{table*}
% To further validate our approach’s effectiveness, we compare it with three diverse long-context reasoning data construction baselines (LoongReason\cite{ling2025longreason}, QwenDocQA\cite{wan2025qwenlong}, LoongRL\cite{wang2025loongrl}) on the DeepSeek-R1-distill-32B backbone, which respectively construct scalable ultra-long algorithmic reasoning task data via algorithmic problem collection and input generator design, synthesize high-quality document QA multi-hop reasoning data by disassembling real documents into fact units and adding interference information, and transform short multi-hop QA data into high-difficulty long-context data through interference document insertion and KeyChain-based original question hiding.

\paragraph{Performance Superiority.}
As presented in Table~\ref{tab:main_results}, while existing strategies like QwenDocQA and LoongRL effectively improve the baseline (raising the average score from $46.30\%$ to $51.31\%$ and $50.71\%$, respectively), our method demonstrates superior efficacy. Without relying on external data, \texttt{DeepSeek-R1-distill-32B+Ours} achieves an average score of $\mathbf{54.03\%}$, outperforming the robust LoongRL baseline by $3.32\%$ and the original base model by $7.73\%$. This significant margin indicates that our construction strategy captures critical long-context dependencies more effectively than previous approaches.

\paragraph{Effectiveness on More Backbone Models}

% \paragraph{Effectiveness Across Model Scales.}
To broadly evaluate the effectiveness of our approach, we implement it on top of the \texttt{Qwen2.5} family and the results are shown in Table ~\ref{tab:comparison_results}. For \texttt{Qwen2.5-14B-instruct}, our approach achieves a remarkable performance boost, increasing the average score from $35.59\%$ to $\mathbf{45.83\%}$ (an absolute gain of $+10.24\%$). Notably, on the \textit{Ruler\_qa2} and \textit{Browscomplong} datasets, our method yields absolute gains of over $20\%$ ($40.28\% \rightarrow 63.21\%$ and $47.30\% \rightarrow 69.72\%$, respectively). Similarly, for the larger \texttt{Qwen2.5-32B-instruct}, our method improves the average performance from $41.54\%$ to $\mathbf{45.50\%}$, demonstrating robustness across different parameter scales.

% To further validate the effectiveness of our approach, we compare it with the data construction method of Loongrl \cite{wang2025loongrl}, a robust baseline for enhancing long-context reasoning. We conduct these experiments on the strong DeepSeek-R1-distill-32B backbone.

% To further validate the effectiveness of our approach, we benchmark it against three representative long-context data construction strategies on the DeepSeek-R1-distill-32B backbone: 
% (1) \textbf{LoongReason}~\cite{ling2025longreason}, which constructs scalable ultra-long algorithmic tasks via input generators; 
% (2) \textbf{QwenDocQA}~\cite{wan2025qwenlong}, which synthesizes multi-hop reasoning data by disassembling documents and injecting interference; and 
% (3) \textbf{LoongRL}~\cite{wang2025loongrl}, which transforms short QA into long-context tasks using interference insertion and KeyChain-based question hiding.

% \paragraph{Superiority over Loongrl.}
% As shown Table \ref{tab:comparison_results}, the Loongrl strategy is indeed effective. The \textit{1k Loongrl} and \textit{4k Loongrl} variants improve the baseline average score from $46.30\%$ to $50.56\%$ and $50.71\%$, respectively. However, our method demonstrates superior efficacy. Without relying on external Loongrl data, DeepSeek-R1-distill-32B+Ours achieves an average score of $\mathbf{54.03\%}$, outperforming the best Loongrl variant (4k) by $3.32\%$ and the original baseline by $7.73\%$. This indicates that our method captures critical long-context dependencies that may be overlooked by standard Loongrl construction.

\paragraph{Synergistic Effect.}
We hypothesize that our method and Loongrl data address different aspects of long-context capabilities. Experimental results support this hypothesis: combining our method with \textit{$1,000$ LoongRL} data yields the highest overall performance of $\mathbf{54.46\%}$. This ``stacking'' effect suggests that our method is not merely a replacement but a complementary enhancement that can be integrated with LoongRL data construction pipelines to push the boundaries of long-context understanding.

\subsection{In-depth Analyses}

\subsubsection{Impact of Atomic Capabilities}
\begin{figure}[t]

        \centering
    \includegraphics[width=0.95\linewidth]{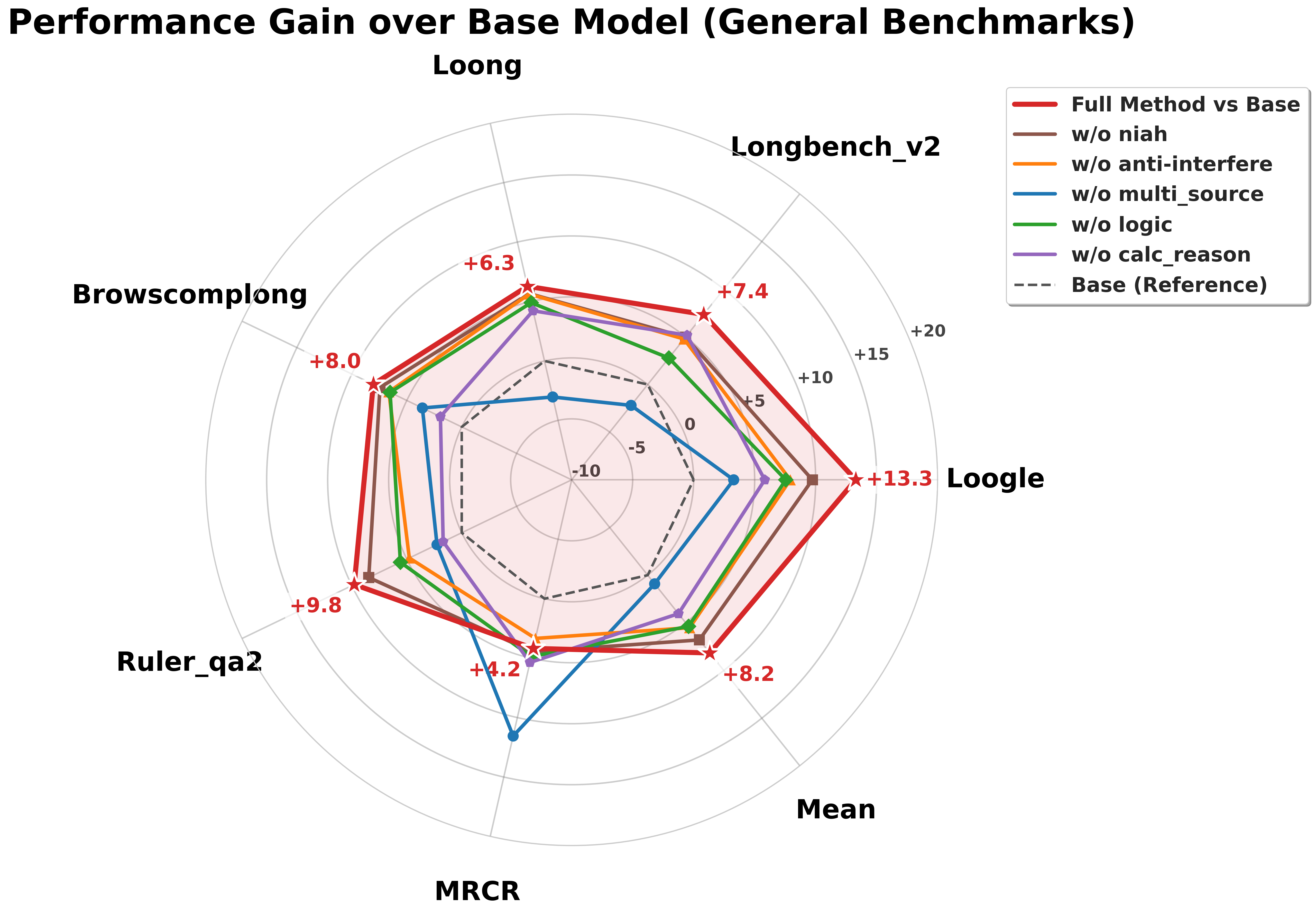}
    \caption{\textbf{Performance Gain over Base Model.} The radar chart compares the performance improvements of our full method (red, with stars) against various ablation variants across six real-world long-context benchmarks.}
    \label{fig:radar_performance}
\end{figure}

To verify the contribution of each atomic capability, we conducted an ablation study by removing specific components from our training data. Figure~\ref{fig:radar_performance} illustrates the performance gains over the Base model (represented by the grey dashed hexagon at 0).

\paragraph{Synergy of Atomic Capabilities.}
\label{radar_analysis}
The \textbf{Full Method} (red line) consistently yields the highest improvements, completely enveloping all other ablated variants in Figure~\ref{fig:radar_performance}. It achieves remarkable gains across diverse benchmarks, such as \textbf{+13.3} on \textit{Loogle} and \textbf{+9.8} on \textit{Ruler-qa2}. This comprehensive superiority demonstrates that the synergy of all atomic capabilities is essential for maximizing robust long-context performance.

\paragraph{Criticality of Multi-source Integration.}
Removing \textbf{Multi-source} data (blue line) reveals a critical phenomenon: performance on general benchmarks like \textit{Loong} and \textit{LongBench-v2} drops below the Base model, despite remaining positive on MRCR. This suggests that Multi-source data acts as a foundational stabilizer, without which the model develops a skill imbalance that degrades its fundamental ability to process general long contexts.

\paragraph{Impact of Logical Reasoning.}
The removal of \textbf{Logic} (green line) leads to substantial performance degradation on \textit{Browscomplong} and \textit{Ruler\_qa2}. The wide gap between the green and red lines on these axes suggests that tasks involving long-document browsing or complex QA rules rely heavily on the model's logical structure and reasoning chain, rather than simple retrieval.

\paragraph{Generalization via Calculation.}
The \textbf{Calc\_reason} capability (purple line) proves to be a global performance enhancer, not limited to the numerical tasks in \textit{Loogle}. The consistent drops across general benchmarks like \textit{Ruler\_qa2} and \textit{Loong} indicate that training on calculation data instills a rigorous reasoning mindset, improving the model's generalized ability to track complex dependencies and maintain precision over long contexts.

\subsubsection{Non-Orthogonality and Hierarchical Dependencies}

\begin{figure}[t]
    \centering
    \includegraphics[width=0.97\linewidth]{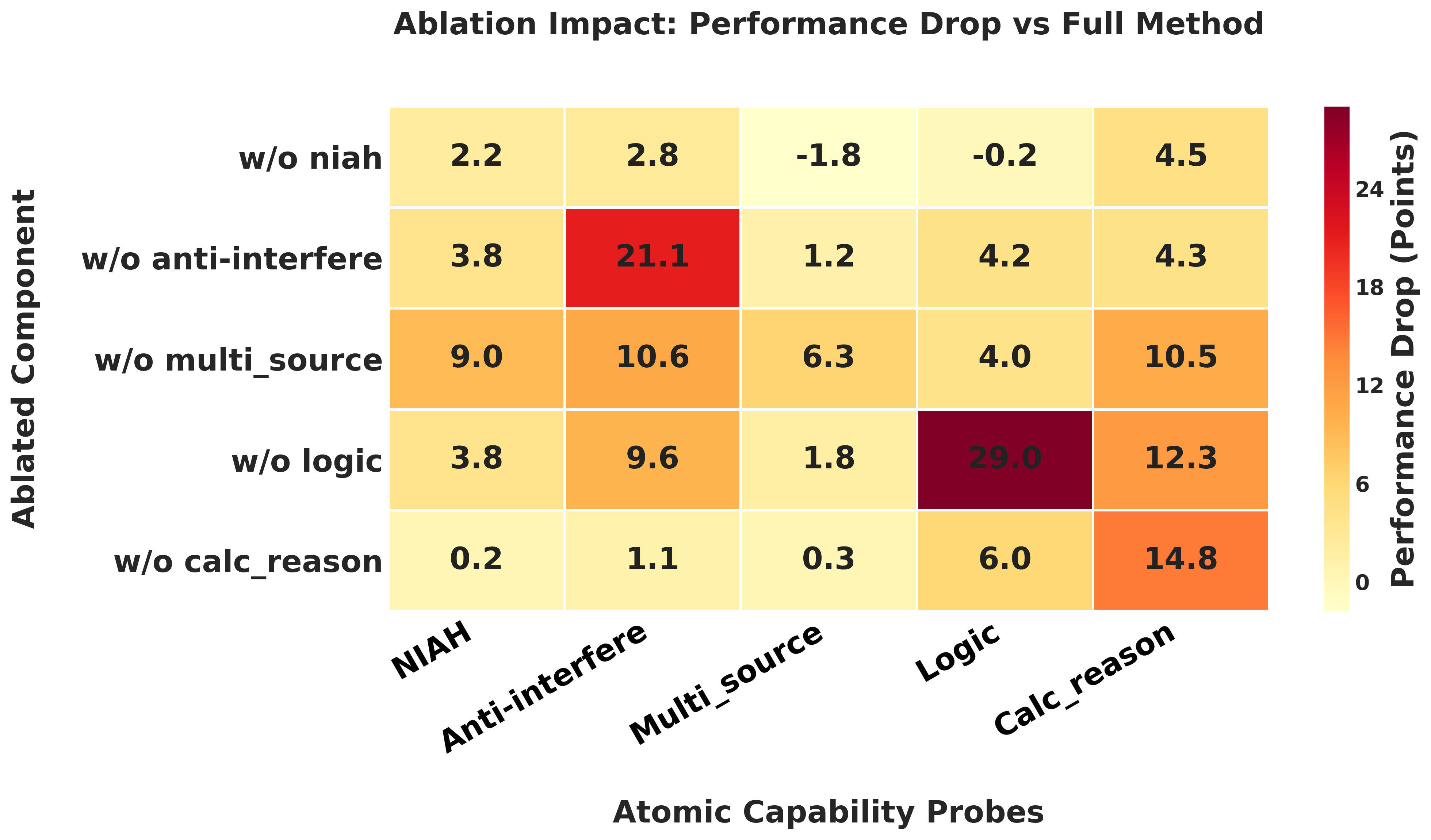}
    \caption{\textbf{Non-Orthogonality Analysis: Performance Drop by Module Removal.} The heatmap illustrates the performance degradation across different atomic capability probes when specific training modules are ablated. }
    \label{fig:non_orthogonality}
\end{figure}

To validate our hypothesis that long-context capability is a hierarchical spectrum rather than a monolithic skill, we analyzed the performance drops on atomic probes when individual training components were ablated (Figure~\ref{fig:non_orthogonality}).

\paragraph{Diagonal Dominance (Distinctness).}
The heatmap exhibits a strong diagonal pattern, particularly for \textbf{Logic} (-29) and \textbf{Anti-interfere} (-21.1). This confirms that "Logical Structuring" and "Robust Discrimination" are specialized skills requiring dedicated training, as they cannot be implicitly learned solely through simple retrieval tasks.

\paragraph{Hierarchical Dependencies.}
The off-diagonal values reveal a clear cognitive hierarchy. We observe an asymmetric dependency where removing \textbf{Logic} significantly impairs \textit{Calc\_reason} (-12.3), whereas removing \textit{Calc\_reason} has a much smaller impact on \textit{Logic} (-6.0). This supports the hypothesis that dynamic state manipulation relies on underlying logical structuring. Furthermore, the removal of \textbf{Multi\_source} causes consistent degradation across all atomic capabilities (e.g., -10.6 on \textit{Anti-interfere}, -10.5 on \textit{Calc\_reason}). This corroborates the "skill imbalance" observed in ~\ref{radar_analysis}, confirming that \textit{Global Integration} acts as a \textbf{foundational stabilizer}—essential for maintaining the general distribution alignment required to support specialized cognitive skills.

\subsubsection{Analysis of Atomic Capability Enhancement}
\begin{figure}
    \centering
    \includegraphics[width=0.97\linewidth]{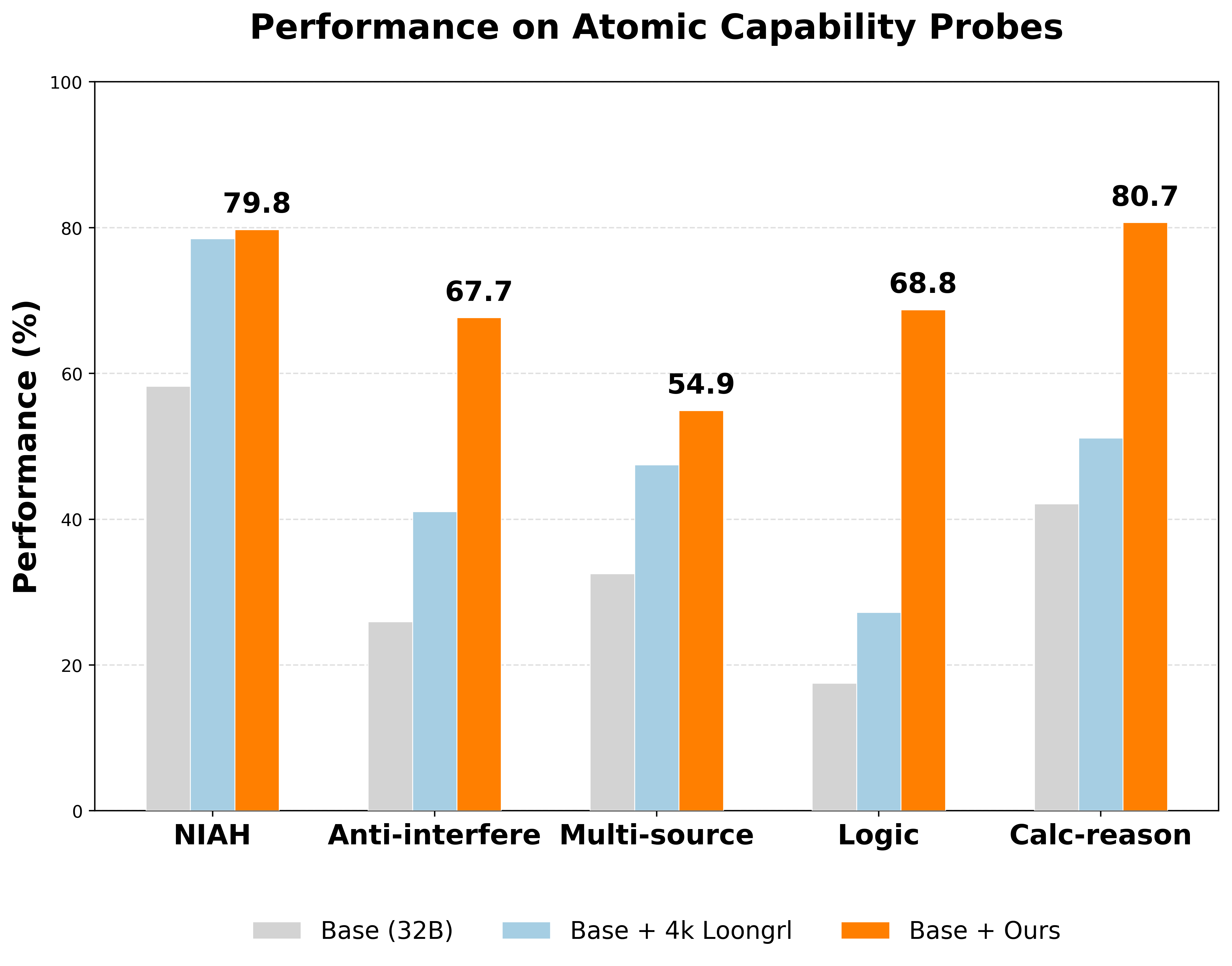}
    \caption{ Performance comparison on Atomic Capability Probes. We compare the DeepSeek-R1-distill-32B base model (Grey), the model trained with 4k LoongRL (Blue), and our proposed method (Orange). }
    \label{fig:atomic_capability_results}
\end{figure}

To evaluate specific capability enhancements, we compared our method against the Base model and the LoongRL baseline (Figure~\ref{fig:atomic_capability_results}).

\textbf{Transformative Gains in Complex Reasoning.}
Our method yields substantial improvements over the Base model across all dimensions, particularly in tasks requiring deep cognitive processing. Notably, \textit{Logic} surges from $\sim18\%$ to \textbf{68.8\%}, and \textit{Calc\_reason} nearly doubles to \textbf{80.7\%}. These results confirm that our approach effectively activates the model's ability to handle complex numerical and logical reasoning within long contexts.

\textbf{Surpassing the Retrieval Ceiling.}
A comparative analysis reveals the limitations of standard data construction. While \textit{LoongRL} matches our performance on simple retrieval (\textit{NIAH}: $\sim78\%$ vs. \textbf{79.8\%}), it fails to generalize to higher-order tasks. Our method significantly outperforms LoongRL on \textit{Anti-interfere} ($+26.7\%$) and \textit{Logic} ($+41.8\%$). This demonstrates that while standard long-context data improves window utilization, our synthesized data is essential for bridging the gap between simple retrieval and complex problem-solving.

\subsubsection{Performance Analysis across Context Length Intervals}
\begin{figure}
    \centering
    \includegraphics[width=0.97\linewidth]{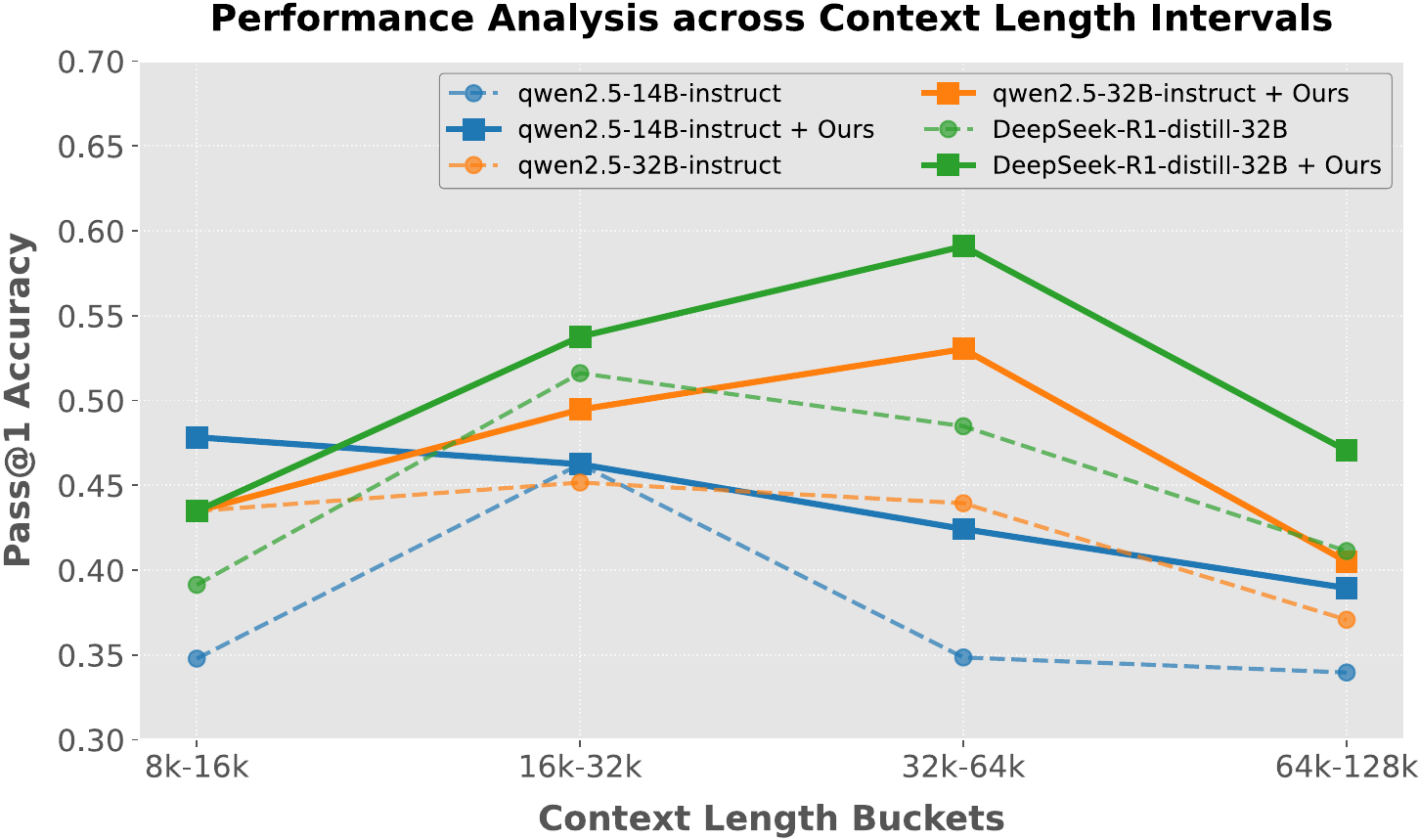}
    \caption{\textbf{Performance Comparison across Context Length Intervals on LongBench-v2.} The Pass@1 accuracy of baseline models (dashed lines) versus our method (solid lines) across different length buckets. }
    \label{fig:length_interval_perf}
\end{figure}
To investigate the robustness of our method under varying input lengths, we analyze performance on LongBench-v2 across four distinct intervals ranging from 8k to 128k. As illustrated in Figure \ref{fig:length_interval_perf}, our method consistently shifts the performance curve upward, maintaining superiority regardless of context length.

\paragraph{Length-Invariant Robustness} The results confirm that our approach effectively mitigates the performance degradation typically observed in extended contexts. Specifically, for the \texttt{DeepSeek-R1-distill-32B} model, we achieve a substantial gain in the 32k-64k interval, boosting accuracy from 48.48\% to \textbf{59.09\%}. Crucially, this advantage persists even in the challenging 64k-128k bucket, demonstrating that our method sustains high-quality reasoning capabilities across the entire long-context spectrum without suffering from significant information loss.

\section{Related Works}
Enhancing the long-context reasoning of LLMs is a crucial yet challenging research problem, attracting extensive efforts~\cite{chen2023longlora,li2024making,bai2024longalign}. Conventional paradigms for enhancing long-context reasoning typically curate task-specific datasets~\cite{chen2023longlora,bai2024longalign} and then optimize LLMs via fine-tuning~\cite{li2024alr,zhang2025longreward} or reinforcement learning~\cite{QwenLongL1,QwenLong1.5,wang2025loongrl}. Nevertheless, long-context reasoning is an inherently complex and monolithic task, making the construction of high-quality training data for this task fraught with intractable challenges ~\cite{yang2025longfaith}, including the misinformation risk caused by inadequate verification protocols in data curation \cite{li2024large} and latent knowledge conflicts existing in the manually or automatically curated datasets ~\cite{xu2024knowledge}. In response to these limitations, we embrace a decomposition perspective and propose AbR framework that breaks down long-context reasoning into atomic skills. This approach enables the automatic curation of verifiable training data, effectively mitigating the data quality and scalability bottlenecks inherent in conventional monolithic paradigms. In parallel, we also examine the impact of table-style tasks on long-context reasoning in another work.
% we embrace a decomposition perspective to break down long-context reasoning into five core atomic skills and propose an anchor-based automatic pipeline integrated with intrinsic verification mechanisms for curating high-quality training data for each modular atomic task; this design not only directly mitigates the intractable data construction challenges plaguing traditional monolithic long-context reasoning research, but also endows our approach with rigorous verifiability and strong scalability, thus providing a targeted, generalizable solution for enhancing LLMs’ long-context reasoning capabilities.

%#########################
\section{Conclusions}
This paper presents a decomposition perspective to long-context reasoning for LLMs and decomposes the long-context reasoning capability into five atomic skills. Then it designs an automatic pipeline to curate training data for each of these skills. Empirical experiments demonstrate that these atomic skills correlate well with standard long-context reasoning benchmarks. Based on this finding, it proposes an effective approach based on reinforcement learning to train LLMs on the curated atomic dataset, in the hope of enhancing the long-context reasoning capability. Intensive experiments on six standard long-context reasoning benchmarks indeed show that the proposed approach yields a substantial gain over the strong backbone LLMs and outperforms several baselines on long-context reasoning. 
\nocite{langley00}

\section*{Impact Statement}
This paper presents work whose goal is to advance the field of Machine Learning. There are many potential societal consequences of our work, none of which we feel must be specifically highlighted here.
% Authors are \textbf{required} to include a statement of the potential broader
% impact of their work, including its ethical aspects and future societal
% consequences. This statement should be in an unnumbered section at the end of
% the paper (co-located with Acknowledgements -- the two may appear in either
% order, but both must be before References), and does not count toward the paper
% page limit. In many cases, where the ethical impacts and expected societal
% implications are those that are well established when advancing the field of
% Machine Learning, substantial discussion is not required, and a simple
% statement such as the following will suffice:

% ``This paper presents work whose goal is to advance the field of Machine
% Learning. There are many potential societal consequences of our work, none
% which we feel must be specifically highlighted here.''

% The above statement can be used verbatim in such cases, but we encourage
% authors to think about whether there is content which does warrant further
% discussion, as this statement will be apparent if the paper is later flagged
% for ethics review.

\bibliography{example_paper}
\bibliographystyle{icml2026}

%%%%%%%%%%%%%%%%%%%%%%%%%%%%%%%%%%%%%%%%%%%%%%%%%%%%%%%%%%%%%%%%%%%%%%%%%%%%%%%
%%%%%%%%%%%%%%%%%%%%%%%%%%%%%%%%%%%%%%%%%%%%%%%%%%%%%%%%%%%%%%%%%%%%%%%%%%%%%%%
% APPENDIX
%%%%%%%%%%%%%%%%%%%%%%%%%%%%%%%%%%%%%%%%%%%%%%%%%%%%%%%%%%%%%%%%%%%%%%%%%%%%%%%
%%%%%%%%%%%%%%%%%%%%%%%%%%%%%%%%%%%%%%%%%%%%%%%%%%%%%%%%%%%%%%%%%%%%%%%%%%%%%%%
% \newpage
% \appendix
% \onecolumn

% You can have as much text here as you want. The main body must be at most $8$
% pages long. For the final version, one more page can be added. If you want, you
% can use an appendix like this one.

% The $\mathtt{\backslash onecolumn}$ command above can be kept in place if you
% prefer a one-column appendix, or can be removed if you prefer a two-column
% appendix.  Apart from this possible change, the style (font size, spacing,
% margins, page numbering, etc.) should be kept the same as the main body.

\newpage
\onecolumn % Remove this if you prefer a two-column appendix
\appendix

\section{Showcases for 5 Atomic Skills}
\label{showcases}
\subsection{Foundational Retrieval: NIAH}
Multiple specific anchor-question pairs are distributed across a long context. The model is tested on its ability to precisely locate a specific anchor and other similar anchors, and answer associated objective questions.
\begin{center}
    \fbox{
    \begin{minipage}{0.95\textwidth}
        \textbf{Case ID:} Distributed-Retrieval-NIAH \\
        \textbf{Category:} NIAH \\
        \textbf{Key Mechanism:} \textit{Distributed Anchors}
        
        \vspace{0.2cm}
        \hrule
        \vspace{0.2cm}
        
        \textbf{Context Overview:}
        \begin{itemize}
            \item \textbf{The Haystack (Background Context):} A long sequence of unrelated text segments (e.g., financial reports, historical essays, or technical logs) serving as noise.
            
            \item \textbf{Inserted Needles (Distributed Pairs):} Specific Anchor-Question pairs are inserted at random intervals throughout the context.
            \begin{quote}
                \small
                \textit{[Segment 0-10\%]} ... The industrial revolution marked a turning point ... \textbf{[ID: A-105] $\rightarrow$ \{Question 1\}} ... \\
                \textit{[Segment 40-50\%]} ... regarding the molecular structure of polymers ... \textbf{[ID: B-292] $\rightarrow$ \{Question 2\}} ... \\
                \textit{[Segment 80-90\%]} ... market volatility observed in the last quarter ... \textbf{[ID: C-345] $\rightarrow$ \{Question 3\}} ...
            \end{quote}
            
            \item \textbf{Target Needle:} The specific pair required by the instruction (e.g., the pair located at the 50\% depth).
        \end{itemize}
        
        \vspace{0.1cm}
        \hrule
        \vspace{0.2cm}
        
        \textbf{Instruction:} \\
        Please answer the question following the anchor \texttt{[ID: B-292]} in the text above.
        
        \vspace{0.2cm}
        \textbf{Target Answer:} \\
        \texttt{3398} \textit{(The model must retrieve the exact objective question associated with the specific key, and answer the question correctly.}
        
    \end{minipage}
    }
    
    \label{fig:niah_retrieval_case}
\end{center}
\subsection{Robustness to Noise: Anti-Interference}
The anchor string 1760536321726-5em0 appears in Doc 1 (twice) and Doc 3 (once). The prompt explicitly requests the content from Doc 3. Tests if the model respects document boundaries despite conflicted retrieval anchors.

\begin{figure}[H]
    \centering
    \fbox{
    \begin{minipage}{0.95\textwidth}
        \textbf{Case ID:} Anchor-Conflict \\
        \textbf{Category:} Anti-interfere \\
        \textbf{Key Mechanism:} \textit{Anchor Duplication \& Document Scope Constraint}
        
        \vspace{0.2cm}
        \hrule
        \vspace{0.2cm}
        
        \textbf{Context Overview:}
        \begin{itemize}
            \item \textbf{Document 1 (Distractor):} Contains the anchor string twice, associated with a Geometry problem and a Matrix Trace problem.
            \begin{quote}
                \small
                ... [Noise] ... \textbf{1760536321726-5em0}: Given a set of points \(P=\{P_1, \dots\}\) on a plane... [Noise] ... \textbf{1760536321726-5em0}: Calculate the trace of the following matrix...
            \end{quote}
            
            \item \textbf{Document 2 (Noise):} Irrelevant content (Programming Q\&A).
            
            \item \textbf{Document 3 (Target):} Contains the anchor string once, associated with a Complex Analysis problem.
            \begin{quote}
                \small
                ... [Noise] ... \textbf{1760536321726-5em0}: Evaluate the integral \( \int _C \frac{\exp(z^2)}{z^2\left(z-1-i\right)}dz \), where \( C \) consists of the contour...
            \end{quote}
        \end{itemize}
        
        \vspace{0.1cm}
        \hrule
        \vspace{0.2cm}
        
        \textbf{Instruction:} \\
        Please answer the question following `\texttt{1760536321726-5em0}' in \textbf{document 3}.
        
        \vspace{0.2cm}
        \textbf{Target Answer:} \\
        \(\pi e^{2i}\) \textit{(The model must ignore the questions in Document 1 and solve the integral in Document 3).}
        
    \end{minipage}
    }
    \label{fig:anchor_conflict_case}
\end{figure}

\subsection{Global Integration: Multi-Source Information Processing}
A single mathematical problem is split into three parts (Setup, Question 1, Question 2) across three different documents. The model must perform cross-document retrieval to reconstruct the full problem context before solving it.
\begin{center}
    \centering
    \fbox{
    \begin{minipage}{0.95\textwidth}
        \textbf{Case ID:} Global-Integration \\
        \textbf{Category:} Global Integration \\
        \textbf{Key Mechanism:} \textit{Fragmented Information Aggregation}
        
        \vspace{0.2cm}
        \hrule
        \vspace{0.2cm}
        
        \textbf{Context Overview:}
        \begin{itemize}
            \item \textbf{Document 1 (Problem Setup):} Contains the initial conditions of the geometry problem embedded within medical text.
            \begin{quote}
                \small
                ... [Medical Q\&A Noise] ... \textbf{h6qKmUOmz2m}: Given circle \(C: x^{2}+y^{2}+2x-2y-6=0\), line \(l\) passes through point \(P(1, 2)\) and intersects circle \(C\) at points \(A\) and \(B\). ... [Noise] ...
            \end{quote}
            
            \item \textbf{Document 2 (Sub-question 1):} Contains the first part of the specific question embedded within different medical/pharmaceutical text.
            \begin{quote}
                \small
                ... [Medical Q\&A Noise] ... \textbf{h6qKmUOmz2m}: (1) If \(\triangle ABC\) is an isosceles right-angled triangle, find the equation of line \(l\); ... [Noise] ...
            \end{quote}
            
            \item \textbf{Document 3 (Sub-question 2):} Contains the second part of the question embedded within a financial project report.
            \begin{quote}
                \small
                ... [Financial Report Noise] ... \textbf{h6qKmUOmz2m}: (2) When \(PC \perp l\), find the equation of the circumcircle of \(\triangle ABC\). ... [Noise] ...
            \end{quote}
        \end{itemize}
        
        \vspace{0.1cm}
        \hrule
        \vspace{0.2cm}
        
        \textbf{Instruction:} \\
        Please assemble the question corresponding to `\texttt{h6qKmUOmz2m}' and answer it.
        
        \vspace{0.2cm}
        \textbf{Target Answer:} \\
        \textbf{(1)} \(x=1\) or \(3x+4y-11=0\) \\
        \textbf{(2)} \(5x^2+5y^2-6x-18y+2=0\) 
        
        \vspace{0.1cm}
        \textit{(The model must retrieve the setup from Doc 1, combine it with conditions from Doc 2 and Doc 3, and solve the aggregated geometry problem).}
        
    \end{minipage}
    }
    
    \label{fig:global_integration_case}
\end{center}

\subsection{Relational Reasoning: Structure Understanding and Logic}
The model must perform a global scan to determine anchor uniqueness frequencies, select the correct document based on key density, and apply strict positional logic to locate the target question, filtering out "trap" anchors (duplicates) during the process.
\begin{center}
    \centering
    \fbox{
    \begin{minipage}{0.95\textwidth}
        \textbf{Case ID:} Logic-Constrained \\
        \textbf{Category:} Relational Reasoning \\
        \textbf{Key Mechanism:} Global Frequency Analysis \& Relative Positional Logic
        
        \vspace{0.2cm}
        \hrule
        \vspace{0.2cm}
        
        \textbf{Context Overview:}
        \begin{itemize}
            \item \textbf{Document 1 (E-Commerce Report):} Contains multiple unique keys associated with Math and Logic problems.
            \begin{quote}
                \small
                ... Consumer Expectations in 2018 ... \textbf{OSXVANVP}: Sequence problem \(a_{n+1}=a_{n}+2^{n}\) ... \textbf{BEKAHJXW}: Statistics problem ...
            \end{quote}
            
            \item \textbf{Document 2 (Environmental Report):} Plain text with no embedded keys (Distractor).
            
            \item \textbf{Document 3 (Biography):} Contains a mix of unique keys and a \textit{duplicated} key.
            \begin{quote}
                \small
                ... General Walker commanded the Eighth Army ... \textbf{KNGUKM}: Tennis tournament logic ... \textbf{ABNKRKRH}: Inequality problem \(\frac{\ln a}{e^a} = \dots\) ... \textbf{GIEDWE}: Physics wave calculation ... \textbf{RJTGAYG}: House logic puzzle ...
            \end{quote}
            
            \item \textbf{Document 4 (Linear Algebra):} Contains the \textit{duplicated} key found in Document 3.
            \begin{quote}
                \small
                ... Problem 5: Decide if range of map ... \textbf{GIEDWE}: Geometry point set problem ...
            \end{quote}
        \end{itemize}
        
        \vspace{0.1cm}
        \hrule
        \vspace{0.2cm}
        
        \textbf{Instruction:} \\
        First, identify anchors that appear \textbf{only once} across all documents. Find the document with the highest total count of anchors. In that document, locate the \textbf{last unique anchor} and answer the question associated with the \textbf{unique anchor immediately preceding it}.
        
        \vspace{0.2cm}
        \textbf{Target Answer (for ABNKRKRH):} \\
        \textbf{C} \textit{(Based on the analysis of the inequality \(\frac{\ln a}{e^{a}}=\frac{\ln b}{b}=-\frac{\ln c}{c}<0\), implying \(a < b < c\)).}
        
    \end{minipage}
    }
    \label{fig:logic_retrieval_case}
\end{center}

\subsection{Dynamic State Tracking: Long-Range Computational Reasoning}
The model cannot simply retrieve a value; it must first determine the state of the context (counting specific key occurrences), evaluate a logical condition based on that state, and then perform a specific sequence of mathematical operations on values retrieved from distributed anchors.
\begin{center}
    \centering
    \fbox{
    \begin{minipage}{0.95\textwidth}
        \textbf{Case ID:} Dynamic-State-Tracking-Math \\
        \textbf{Category:} Dynamic State Tracking / Computational Reasoning \\
        \textbf{Key Mechanism:} \textit{Conditional Logic \& Multi-Stage Aggregation}
        
        \vspace{0.2cm}
        \hrule
        \vspace{0.2cm}
        
        \textbf{Context Overview:}
        \begin{itemize}
            \item \textbf{Document 1 (Survey Analysis):} Contains a key embedded in statistical text.
            \begin{quote}
                \small
                ... excluding incomplete questionnaires ... \textbf{LTUCRHGAXK}: \(82 \times 67\) ...
            \end{quote}
            
            \item \textbf{Document 2 (Game Design Manual):} Contains a key embedded in UI/UX instructions.
            \begin{quote}
                \small
                ... visual balance lines ... \textbf{NNQABR}: \(6 - (8 - 4) + 16\) ...
            \end{quote}
            
            \item \textbf{Document 3 (Stock Market Report):} Contains the target keys for the "False" branch.
            \begin{quote}
                \small
                ... A-share defense battle ... \textbf{ROXBXXYMYU}: \(2 - 14.9 / 6 + 7\) ... market leverage ... \textbf{MPEOEZHO}: \(19 \times 19.7 + 20 / 2\) ...
            \end{quote}
            
            \item \textbf{Document 4 (Electricity Report):} Contains the target keys for the "True" branch (Distractors).
            \begin{quote}
                \small
                ... 16-candle lamps installed ... \textbf{TBWYKIE}: \((20 \times 4) \times (3 - 16)\) ... \textbf{KYPNFYC}: \(\sin(135^\circ) \times \cos(135^\circ) + 3\) ...
            \end{quote}
        \end{itemize}
        
        \vspace{0.1cm}
        \hrule
        \vspace{0.2cm}
        
        \textbf{Instruction:} \\
        If the number of documents containing \texttt{LTUCRHGAXK} is greater than the number of documents containing \texttt{NNQABR}, calculate the sum of the answers for \texttt{TBWYKIE} and \texttt{KYPNFYC}. Otherwise, calculate the product of the answers for \texttt{ROXBXXYMYU} and \texttt{MPEOEZHO}. (Round intermediate steps to 2 decimal places).
        
        \vspace{0.2cm}
        \textbf{Target Answer:} \\
        \textbf{2505.64}
        
    \end{minipage}
    }
    
    \label{fig:dynamic_state_tracking}
\end{center}

\section{Training Details}
\label{training_details}

This section details the reinforcement learning algorithms and techniques utilized for enhancing atomic skills, including the optimization objective, sampling strategies, and reward mechanisms.

\subsection{Group Relative Policy Optimization (GRPO)}

Group Relative Policy Optimization (GRPO)~\cite{shao2024deepseekmath} is designed to reduce the computational overhead in large-scale model training by eschewing a trained value function. The core innovation lies in its advantage estimation mechanism.

Instead of relying on a critic network to provide a value baseline, GRPO computes the advantage for each sample relative to the average reward of a group of samples generated from the same prompt. For a given prompt, a group of $\mathcal{G}$ outputs $\{o_1, o_2, \dots, o_{\mathcal{G}}\}$ is sampled from the policy $\pi_{\text{old}}$. Upon obtaining a reward $r_i$ for each output $o_i$, the advantage $A_i$ is calculated as:
\begin{equation}
    A_i = r_i - \frac{1}{|\mathcal{G}|}\sum_{j=1}^{|\mathcal{G}|} r_j 
    \label{eq:app_grpo_advantage}
\end{equation}
This formulation utilizes the group mean reward as a dynamic baseline. The policy is updated using a clipped surrogate objective augmented with a KL-divergence penalty. Formally, the objective is defined as follows.
\begin{equation}
    \mathbb{E} \left [ \min \Big ( \frac{\pi_{\theta}(a|s)}{\pi_{\text{old}}(a|s)}A_i, 
    \text{clip}\big(\frac{\pi_{\theta}(a|s)}{\pi_{\text{old}}(a|s)}, 1-\epsilon, 1+\epsilon\big)A_i \Big ) \right ] - 
    \beta \cdot D_{\text{KL}}(\pi_{\theta} \| \pi_{\text{old}})
    \label{eq:app_grpo_objective}
\end{equation}

where $\theta$ represents the parameters, $\epsilon$ is the clipping hyperparameter, and $\beta$ controls the strength of the KL regularization.

\subsection{Dynamic Sampling}

To address the issue of reward homogenization—where similar rewards within a group lead to near-zero advantages and vanishing gradients—Dynamic Sampling~\cite{yu2025dapo} is employed. This strategy dynamically prunes samples with redundant rewards during training. By ensuring that training batches are composed of diverse and informative trajectories, this method strengthens the gradient signal and accelerates convergence.

\subsection{Reward Modeling and Reasoning Induction}

The reward signal is typically derived from an LLM-as-a-Judge paradigm, where a larger model evaluates the correctness of generated outputs. A binary reward (1 for a match, 0 otherwise) is assigned based on whether the output matches the reference answer.

Different strategies are applied depending on the model type:
\begin{itemize}
    \item \textbf{Instruct Models:} To guide standard instruction-tuned models toward a deliberate reasoning mode, a specialized ``chain-of-thought'' system prompt is introduced alongside a format-compliance reward signal.
    \item \textbf{Models with CoT:} For models that inherently incorporate a reasoning process, standard training procedures are followed without additional reasoning-inducing prompts.
\end{itemize}

% \subsection{Data Construction Strategy}
% The construction of the training dataset involves a rejection sampling approach. Responses are sampled from a strong teacher model (e.g., DeepSeek V3.1) for each query. To ensure an appropriate level of difficulty for the student model, queries are filtered to retain those exhibiting a specific pass rate range (e.g., between 0.3 and 0.6). The final dataset is typically balanced across different atomic capabilities such as Anti-interfere, Multi-hop, Multi-source, Logic, Calc-reason, and NIAH.
% %%%%%%%%%%%%%%%%%%%%%%%%%%%%%%%%%%%%%%%%%%%%%%%%%%%%%%%%%%%%%%%%%%%%%%%%%%%%%%%
% %%%%%%%%%%%%%%%%%%%%%%%%%%%%%%%%%%%%%%%%%%%%%%%%%%%%%%%%%%%%%%%%%%%%%%%%%%%%%%%

\end{document}